\documentclass{article} 

\usepackage{arxiv} 

\usepackage{authblk}

\usepackage[T1]{fontenc}
\usepackage[utf8]{inputenc}
\usepackage{microtype} 
\usepackage[pdftex]{graphicx} 
\graphicspath{ {./images/} }
\usepackage[dvipsnames]{xcolor}
\usepackage[hidelinks]{hyperref} 

\usepackage{geometry}
\usepackage{amsmath, amsfonts, amssymb, mathtools, bm}
\usepackage{array, booktabs}
\usepackage{enumitem}

\usepackage[english]{babel} 

\usepackage{lipsum}

\title{Large Language Models and Book Summarization: Reading or Remembering, Which Is Better?}

\author[1]{\textbf{Tairan Fu}}
\author[2]{\textbf{Javier Conde}}
\author[2]{\textbf{Pedro Reviriego}}
\author[3]{\textbf{Javier Coronado-Blázquez}}
\author[2,4]{\textbf{Nina Melero}}
\author[5]{\textbf{Elena Merino-Gómez}}

\affil[1]{Politecnico di Milano, Italy}
\affil[2]{Universidad Politécnica de Madrid, Spain}
\affil[3]{Telefónica Tech, AI \& Data Unit, Spain}
\affil[4]{New York University, Madrid Campus, Spain}
\affil[5]{Universidad de Valladolid, Spain}

\date{} 

\begin{document}

\maketitle

\begin{abstract}
Summarization is a core task in Natural Language Processing (NLP). Recent advances in Large Language Models (LLMs) and the introduction of large context windows reaching millions of tokens make it possible to process entire books in a single prompt. At the same time, for well-known books, LLMs can generate summaries based only on internal knowledge acquired during training. This raises several important questions: How do summaries generated from internal memory compare to those derived from the full text? Does prior knowledge influence summaries even when the model is given the book as input? In this work, we conduct an experimental evaluation of book summarization with state-of-the-art LLMs. We compare summaries of well-known books produced using (i) only the internal knowledge of the model and (ii) the full text of the book. The results show that having the full text provides more detailed summaries in general, but some books have better scores for the internal knowledge summaries. This puts into question the capabilities of models to perform summarization of long texts, as information learned during training can outperform summarization of the full text in some cases.
\end{abstract}




\section{Introduction}

Summarization is one of the most common tasks in Natural Language Processing (NLP) \cite{Mani2001}. Over the years, many techniques and models have been used for summarization, typically trained on pairs of documents, summaries. For example, early methods relied on statistical features such as term frequency, sentence position, and cue phrases to select relevant sentences \cite{zhang2024comprehensive}. Later, neural network–based approaches, particularly sequence-to-sequence models, performed summarization by generating new sentences that capture the meaning of the source document \cite{nallapati-etal-2016-abstractive}. More recently, transformer-based pre-trained encoders like BERT have been used to improve summarization performance by using rich contextual representations \cite{liu-lapata-2019-text}. 

The development of Large Language Models (LLMs) opened new directions for summarization, since the models can be given the text and asked to summarize it in a single prompt \cite{basyal2023text}. The use of LLMs can be more sophisticated by using prompt engineering, model fine-tuning, or knowledge distillation to achieve better performance or use smaller models \cite{zhang2024comprehensive}. For well-known texts, another option is to rely on the model's internal knowledge to provide a summary, for example, of a book without providing the text \cite{coronado2025evaluating}.

In fact, LLMs are commonly seen to have two types of memories: internal (or parametric) and external \cite{carlini2022quantifying}. The internal memory is encoded in the model parameters, representing the knowledge acquired during training. This parametric memory has been shown to store facts, linguistic patterns, and in some cases even verbatim passages from training data \cite{morris2025much}. In contrast, the external memory is provided through the context window at inference time. In the case of summarization, it allows one to provide the model with the text to summarize. The size of the context window limits the amount of external memory, with the first generation of models having windows of just a few thousand tokens, while newer models such as GPT-4.1 and Gemini 2.5 reach the million-token range. Since most books have at most a few hundred thousand words, one million tokens is sufficient to pass the entire book in the input prompt for summarization. These two forms of memory can interact with parametric knowledge overriding contextual input, or the two may reinforce each other \cite{cheng2024interplay}. However, the interaction between memorization in weights and processing of contextual input remains an open research question.

In this context, the study of book summarization based on internal and external memory is an interesting topic that can improve our understanding of the inner workings of LLMs and also of their limitations when performing summarization of large texts. This paper takes an initial step in this direction by performing a comparison of book summaries generated by LLMs based on internal and external knowledge with high-quality human summaries. In particular, the following contributions are made:

\begin{enumerate}[nosep]
    \item Conduct to the best of our knowledge the first evaluation comparing LLM generated book summaries with internal and external memory using state-of-the-art models with large context windows.
    \item Create an open repository with the code, books, and summaries of the experiments to facilitate further research on the topic and reproducibility of our results.  
    \item Show that for most books, the use of the full text for summarization improves performance compared to summaries based only on parametric memory.
    \item Show that for some books, internal knowledge summaries get better scores than summaries done with access to the full text of the book.
    \item Discuss the potential reasons for the better performance of internal summaries. 
\end{enumerate}

The rest of the paper is organized as follows. Section \ref{sec:related-work} summarizes related works on text summarization focusing on the use of LLMs and on the different types of memory in LLMs and their interactions. The methodology used for the evaluation is described in Section \ref{sec:methodology} and the results are presented and discussed in Section \ref{sec:results}. The paper ends with the conclusion in Section \ref{sec_conclusion}.

\section{Related work}
\label{sec:related-work}

There are many recent works on summarization and LLM memory. The following subsections provide an overview of current research activities on these topics as well as some historical context. 

\subsection{Text summarization}

As discussed in the Introduction, text summarization has been a central task in NLP for decades \cite{Mani2001}. The techniques used to produce summaries have evolved from early statistical-based methods that rely on word frequencies \cite{luhn1958automatic} to the use of complex neural models such as LLMs. There are two main types of automatically generated summaries: extractive and abstractive \cite{zhang2024comprehensive}. In the first case, the summary is built by extracting fragments from the original text, while in the latter, the process generates new text to summarize the input. Initially, summarization approaches tried to find the most relevant parts of the text by computing metrics on words and sentences and extracting them to build the summary. The use of classical machine learning algorithms was also explored for extractive summarization \cite{chali2009svm}. 

The development of powerful deep learning models led to the use of more advanced neural architectures such as recurrent neural networks (RNN), convolutional neural networks (CNN), and long-short-term memory networks (LSTM) for summarization \cite{nallapati-etal-2016-abstractive},\cite{song2019LSTM}. The introduction of the Transformer architecture revolutionized NLP in general and text summarization in particular. Bidirectional Encoder Representations from Transformers (BERT), with its bidirectional encoder, has been widely used in extractive summarization, where embeddings are leveraged to select the most salient content from a document \cite{liu-lapata-2019-text}. Instead, the Text-to-Text Transfer Transformer (T5) has been used for abstractive summarization, as autoregressive decoders enable the generation of fluent and coherent summaries beyond simple sentence extraction \cite{JMLR:T5}. The use of Transformers set a new state-of-the-art and laid the foundations for the emergence of large language models designed to handle summarization as part of a broader range of complex natural language tasks.

Large Language Models (LLMs) have shown good performance for text summarization, leveraging their extensive training on massive amounts of text to generate coherent, contextually rich, and human-like summaries \cite{zhang2024benchmarkingLLMsummaries}. Unlike earlier task-specific architectures, these models demonstrate strong zero-shot and few-shot capabilities, enabling effective summarization without the need for extensive fine-tuning. Recent benchmarks consistently highlight their superiority over traditional encoder–decoder models, although challenges remain in ensuring factual consistency, mitigating hallucinations, and aligning outputs with user needs \cite{chen2025cothssum},\cite{askari2024assessingllmsummaries}. Early approaches to book summarization were constrained by narrow context windows, necessitating ``divide-and-conquer'' strategies or hierarchical merging of chapter-level summaries \cite{wu2021recursively}. However, a few recent models have introduced context windows spanning millions of tokens, enabling the processing of entire literary works in a single prompt.

\subsection{Large Language Model Memory}

The memory of LLMs has been widely studied to try to understand how models learn and memorize information in their parameters during training \cite{carlini2022quantifying}. In some cases, models reproduce the content of their training data verbatim, raising concerns about data leakage and potential privacy violations \cite{carlini2021extracting}. The amount of information stored by the models appears to be related to the number of parameters, with some works indicating that the information that can be memorized is the number of parameters in the model multiplied by a small number of bits \cite{morris2025much}. 

In addition to this parametric memory, which is filled during model training, LLMs have a dynamic memory that can be defined at inference time and is given by their context window. This window enables the user to provide information, such as text to summarize and the instructions to do it, at inference time. This information is upper-bounded by the maximum input length supported by the model architecture, which constrains the amount of information that can be processed at once. The size of the context window has increased from a few thousand tokens in early models to over one million tokens in state-of-the-art models \cite{team2024gemini}. However, having a large context window does not guaranty that the model can always use this information correctly \cite{jin2024llmcontext}. Several issues have been identified, with models often overemphasizing recent tokens while neglecting information from earlier parts of the input, a phenomenon known as ``Lost in the Middle'' \cite{liu2023lost}. Also, as the input length grows, models also struggle with context fragmentation making it difficult to integrate relevant information consistently across distant segments of the input text \cite{bai-etal-2024-longbench}.

Another interesting topic is the interaction between both types, parametric and dynamic, of memory, with recent studies showing that models tend to rely on the information provided in the input text over their internal knowledge \cite{cheng2024interplay}. This is relevant for book summarization, as for well-known books models will have internal knowledge that can interfere with the book text provided as input.

\section{Methodology}
\label{sec:methodology}

This section describes the methodology and experimental setup used in our evaluation\footnote{The code and data including the books and summaries are available at \url{https://github.com/aMa2210/llm-summary-internal-vs-external}}. First, the overall approach is presented, followed by a discussion of the books, LLMs and prompts considered.


\subsection{Overall approach}

The evaluation follows the procedure used in \cite{coronado2025evaluating} to generate summaries based only on the internal knowledge of the LLM. First, an LLM is used to generate the book summaries. In the case of internal knowledge summaries, we use a prompt only with the instructions, while for the external knowledge summaries, the input includes the full text of the book in addition to the instructions. 

\begin{figure*}[t]
    \centering
    \includegraphics[width=.6\textwidth]{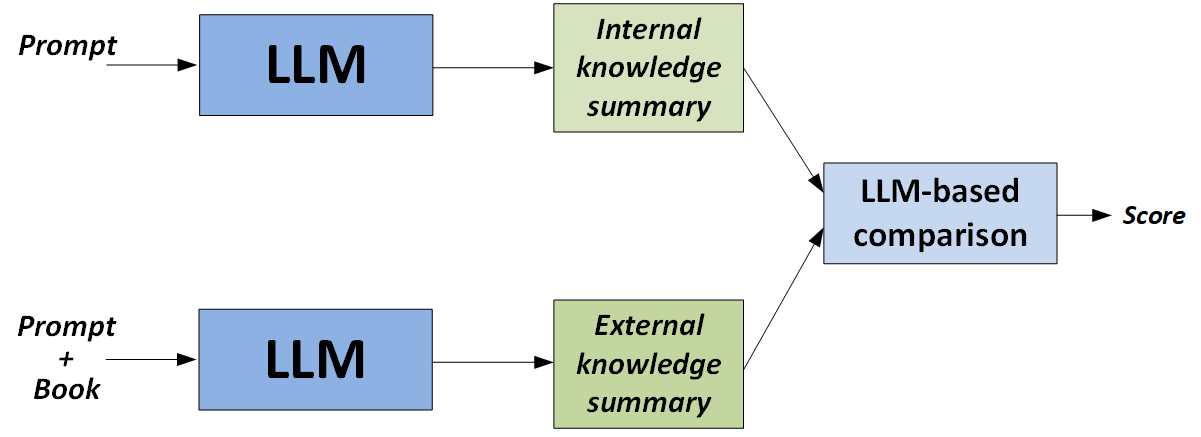}
    \caption{Overall approach to generate and evaluate the summaries.}
    \label{fig:approach}
\end{figure*}

Once the summaries are generated, they are compared. Beyond traditional overlap-based metrics, recent literature has popularized the LLM-as-a-judge framework, in which LLMs are used to evaluate quality across dimensions such as faithfulness, relevance, and coherence, demonstrating high correlation with human judgment \cite{zheng2023judging}. We adopt this scheme, and comparisons are made using the generation model itself or a second LLM. As a result, we have a quality score for each book. The general scheme is illustrated in Figure \ref{fig:approach}.

\subsection{Books}

The selection of books chosen for our evaluation was made to ensure that they are: 1) available in the public domain, 2) widely known, and 3) cover a wide range of authors, genres, and styles. In particular, a subset of the books considered in \cite{coronado2025evaluating} that are in the public domain was used. The selection was made by two experts in literature. The list of books evaluated is shown in Table \ref{tab:authors_books}. It can be seen that all are widely known books that are for sure included in the training datasets of state-of-the-art LLMs and cover different periods and styles. English translations of the original text are used when the book was written in another language. 

There are a number of features that can influence the difficulty of writing a summary for an LLM. For example, intuitively, the longer the book, the harder it is for the LLM to write a summary, as it needs to analyze and relate more tokens. In our set, the longest book, ``The Count of Monte Cristo'', has close to half a million words and uses most of the LLM context window. On the other hand, the shortest one, ``A Doll's House'' has less than thirty thousand words. However, length is not the only feature that affects the difficulty of summarizing. The complexity of the plot, the writing style, or the type of writing: prose, verse, or drama can also make summarizing difficult. For instance, capturing the plot of James Joyce’s Ulysses is intuitively more demanding than that of a conventional novel, and summarizing verse is likewise harder than summarizing prose.





\begin{table*}[htbp]
\scriptsize
\centering
\caption{Books considered in the evaluation}
\begin{tabular}{|c|l|l|}
\hline
\textbf{No.} & \textbf{Author} & \textbf{Book} \\ \hline
1 & Jane Austen & \textit{Pride and Prejudice} \\ \hline
2 & Giovanni Boccaccio & \textit{The Decameron} \\ \hline
3 & Dante Alighieri & \textit{The Divine Comedy} \\ \hline 
4 & Charlotte Brontë & \textit{Jane Eyre} \\ \hline
5 & Lewis Carroll & \textit{Alice's Adventures in Wonderland} \\ \hline
6 & Miguel de Cervantes & \textit{Don Quixote} \\ \hline
7 & Charles Dickens & \textit{Great Expectations} \\ \hline
8 & Daniel Defoe & \textit{Robinson Crusoe} \\ \hline
9 & Fyodor Dostoyevsky & \textit{Crime and Punishment} \\ \hline
10 & Alexandre Dumas & \textit{The Count of Monte Cristo} \\ \hline
11 & F. Scott Fitzgerald & \textit{The Great Gatsby} \\ \hline
12 & Gustave Flaubert & \textit{Madame Bovary} \\ \hline
13 & Nathaniel Hawthorne & \textit{The Scarlet Letter} \\ \hline
14 & Homer & \textit{The Iliad} \\ \hline
15 & Henrik Ibsen & \textit{A Doll's House} \\ \hline
16 & James Joyce & \textit{Ulysses} \\ \hline
17 & Luigi Pirandello & \textit{Six Characters in Search of an Author} \\ \hline
18 & William Shakespeare & \textit{Hamlet} \\ \hline
19 & Mary Shelley & \textit{Frankenstein} \\ \hline
20 & Bram Stoker & \textit{Dracula} \\ \hline
21 & Leo Tolstoy & \textit{Anna Karenina} \\ \hline
22 & Mark Twain & \textit{The Adventures of Huckleberry Finn} \\ \hline
23 & Jules Verne & \textit{20,000 Leagues Under the Sea} \\ \hline
24 & Virginia Woolf & \textit{Mrs. Dalloway} \\ \hline
25 & Thomas Mann & \textit{Death in Venice} \\ \hline
\end{tabular}
\label{tab:authors_books}
\end{table*}



\subsection{LLMs}

The models must be able to store the entire book in the context window to support external knowledge summarization. The longest book considered in the experiments is "The Count of Monte Cristo, which is about half a million words. Since typically up to 1.3-1.4 tokens per word are needed on average, placing the book in the context window can take close to 600,000 tokens\footnote{\url{https://platform.openai.com/docs/concepts/tokens}}. At the time of writing this paper, very few models had context windows capable of storing this amount of information, and most of them were previews or not accessible to all users. There were, however, two widely used models with context windows of one million tokens and state-of-the-art performance: GPT-4.1 and Gemini-2.5, that we selected for our experiments. Summaries were generated with the two models and also evaluated with the two models. This enables us to detect evaluation biases where, for example, a model tends to favor its own summaries \cite{xu2024pride}.   

\subsection{Instructions}

The prompts used to generate the summaries follow the instructions used in \cite{coronado2025evaluating} and focus on providing a detailed account of the plot, events, and characters. For the internal knowledge summaries, the prompt used is as follows: 

\textit{“Provide a very detailed summary of the plot for the book "\{title\}" by \{author\}. The summary must be of the original book, NOT any adaptation like a film or TV show. Include all main events and the complete storyline, detailing every key development, situations, events with characters, and the conclusion. Do NOT include any historical context, literary analysis, or philosophical discussion, only the plot.”}
\newline
\newline

To generate the summaries using the full text of the books, we used a similar prompt, adding that the model should write the summary based on the text provided and not use its internal knowledge:

\textit{“Provide a very detailed summary of the plot for the following book. The summary must be of the text provided, do NOT use any internal or previous
knowledge that you may have on that book to create the summary. Include all main events and the complete storyline, detailing every key development, situations, events with characters, and the conclusion. Do NOT include any historical context, literary analysis, or philosophical discussion, only the plot.
Full text of the book: "\{fulltext of the book\}".}

It can be seen that no information is given on the author or title or the book and the models are instructed not to use any internal knowledge they may have. 

Finally, to compare the LLM-generated summaries, the following prompt was used:

\textit{“Please compare the following two summaries and tell me which is more complete and contains information on the complete storyline, key developments, events or characters missing in the other. If the first contains more information at the end, return a score of 1, if both summaries are the same of 0 and if the second is more complete of -1.
\{LLM generated internal summary\}  
\{LLM generated external summary\}"  
}

As discussed, for this last prompt, use the two models so that GPT-4.1 and Gemini-2.5 judge their own summaries and also those written by the other model. This enables us to detect any potential self-bias in the evaluation \cite{xu2024pride}.

\subsection{Procedure}

As summarization must not introduce modifications to the information provided in the text, initially, we set the temperature to 0.4, a relatively low value. This provides a trade-off between avoiding models from deviating from the text and providing some freedom to write the summaries. To check the impact of the temperature on the summaries, the experiments were also run with a higher temperature value of 0.9 to assess the impact of enabling greater creativity in the models on the quality and variability of the summaries. For each model, we generate the summaries five times as this enables us to evaluate also the consistency of the models when writing summaries. Then comparisons are done pairwise between each of the internal and external summaries, so if the internal summaries are $I_1,I_2,I_3,I_4,I_5$ and the external $E_1,E_2,E_3,E_4,E_5$ we evaluate twenty-five pairs:

\[
\small
\begin{aligned}
\{ &(I_1, E_1), (I_1, E_2), (I_1, E_3), (I_1, E_4), (I_1, E_5), \\
   &(I_2, E_1), (I_2, E_2), (I_2, E_3), (I_2, E_4), (I_2, E_5), \\
   &(I_3, E_1), (I_3, E_2), (I_3, E_3), (I_3, E_4), (I_3, E_5), \\
   &(I_4, E_1), (I_4, E_2), (I_4, E_3), (I_4, E_4), (I_4, E_5), \\
   &(I_5, E_1), (I_5, E_2), (I_5, E_3), (I_5, E_4), (I_5, E_5) \}
\end{aligned}
\]

and add the results to obtain a score for each book in the range of -25 to 25.

\section{Results and Analysis}
\label{sec:results}

\begin{figure}[h]
    \centering
    \includegraphics[width=.7\linewidth]{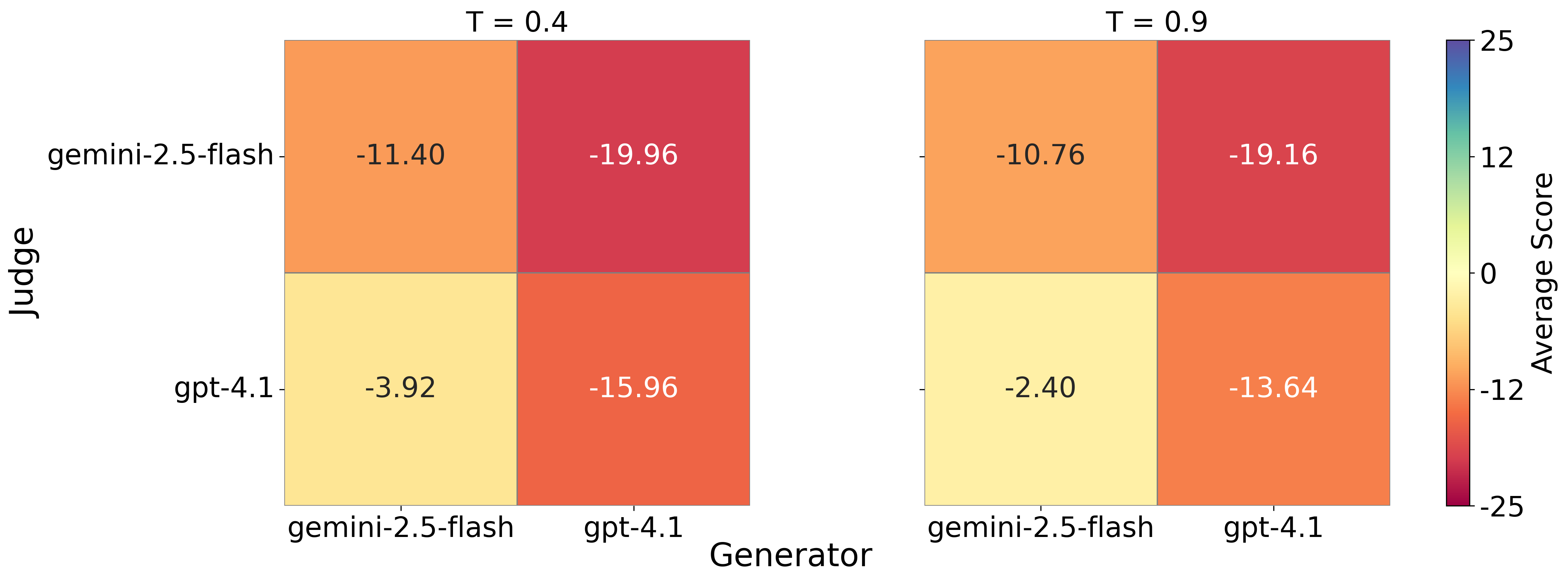}
    \caption{Comparison of the average scores across all the books.}
    \label{fig:all_generators_heatmaps}
\end{figure}

To get an initial idea of the results, Figure \ref{fig:all_generators_heatmaps} shows the average scores of the summaries generated at temperatures 0.4 and 0.9 for the 25 books. For each of them, the generator and judge models are shown horizontally and vertically, resulting in a 2-by-2 matrix for each temperature value.  It can be seen that all values are negative, indicating that external summaries consistently outperform internal ones. Therefore, providing the full text improves the quality of the summation. However, the degree of this improvement depends strongly on both the generator and the judge.

\begin{figure*}[h]
    \centering
    \includegraphics[width=.9\textwidth]{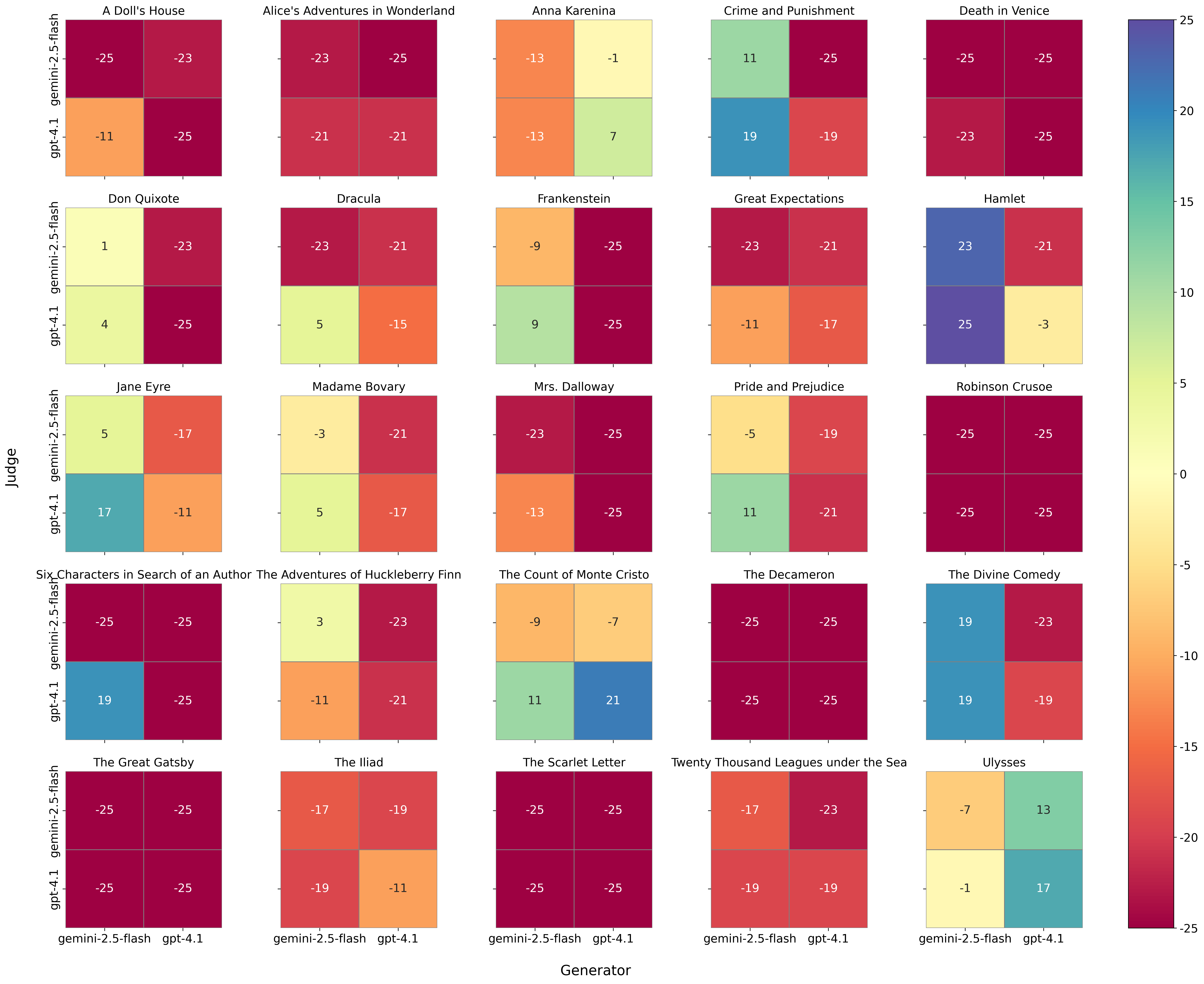}
    \caption{Comparison results obtained at temperature 0.4.}
    \label{fig:compare_t0.4}
\end{figure*}

\begin{figure*}[h]
    \centering
    \includegraphics[width=.9\textwidth]{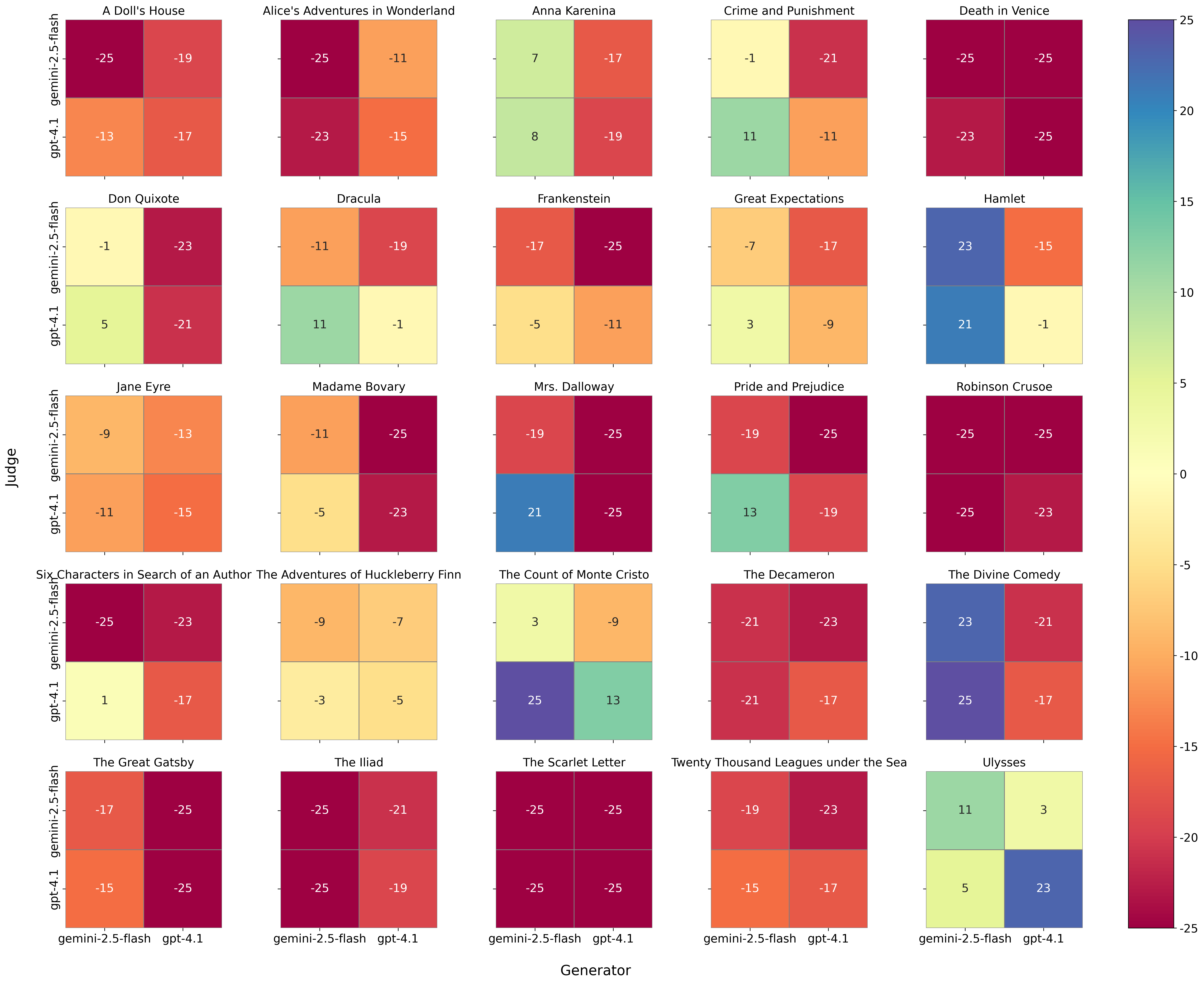}
    \caption{Comparison results obtained at temperature 0.9.}
    \label{fig:compare_t0.9}
\end{figure*}

\begin{figure*}[h]
    \centering
    \includegraphics[width=.9\textwidth]{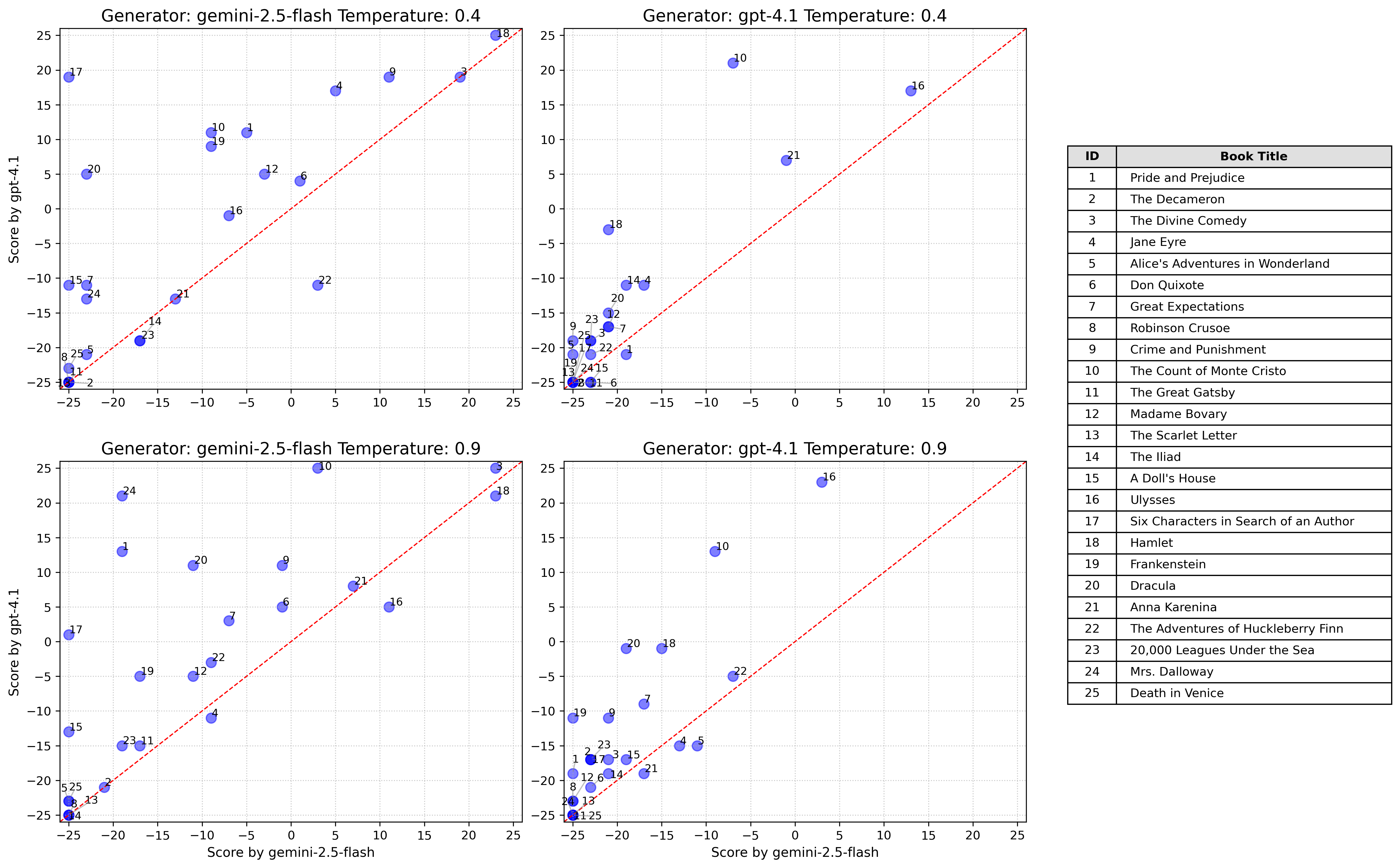}
    \caption{Scatter plots of the summaries scores by evaluator model.}
    \label{fig:delta_comparison}
\end{figure*}

The summaries generated by Gemini-2.5-Flash receive less negative scores regardless of who performs the evaluation. This suggests that Gemini’s is less capable of summarizing from full text or that it maintains a more consistent performance when summarizing from prior knowledge. Regarding the evaluation, GPT-4.1 as a judge tends to produce scores that are significantly higher on average than those of Gemini for the same generator, so favoring internally generated summaries. The temperature has a small impact, increasing the temperature from 0.4 to 0.9 slightly increases the scores, suggesting that higher creativity narrows the gap between internal and external summaries. However, the relative ranking of model and judge combinations remains stable across temperatures. In summary, external context remains decisive, even with extensive prior knowledge, both LLMs produce notably better summaries when given access to the full text.

More detailed information is provided in Figures \ref{fig:compare_t0.4} and 
\ref{fig:compare_t0.9}, which provide the same information but for each of the books. As expected, for most books, scores take negative values in many cases below -20, confirming that external summaries are generally preferred.  However, several books deviate significantly from this pattern, with neutral or positive mean values in one or more generator/judge pairs. The most relevant ones were analyzed in more detail.

\begin{itemize}[nosep]
    \item \textit{The Divine Comedy and Hamlet} had positive results when Gemini-2.5-Flash was the generator (19 to 25) and much lower when GPT4-1 generated the summaries (–3 to –25). 
    \item \textit{Ulysses} had positive results in most cases and large variations when increasing the temperature. 
    \item \textit{The Count of Montecristo} had very high scores, above 20 for some generator/evaluator pairs. 
\end{itemize}

Some samples of the generated summaries for these four books were then manually analyzed. The human evaluation was generally in line with the LLM judgments. In the process, it was found that Gemini's summaries for the Divine Comedy were in some cases incomplete. This was not due to any limitation in the number of output tokens. It may be related to the fact that Gemini's external summaries tend to follow a chapter by chapter structure and the Divine Comedy has a large number of them. This failure illustrates the limitations of LLMs and the need for validation and cross checking of their outputs. These results suggest that for experimental narratives (Ulysses), high variance reflects the struggle to map non-linear structures while for massive plot-driven works (The Count of Monte Cristo), internal summaries avoid the ``lost-in-the-middle'' pitfalls of full-text processing.

This analysis of the outliers reveals that LLM summarization accuracy is text-dependent, not uniformly distributed across the literary corpus, even if all 25 books are classical and very well-known texts, to which models have certainly been extensively exposed during training. Factors such as genre, length, and plot complexity appear to influence performance, with poetry, theater, and long or complex books showing relatively better results for summaries generated from the model’s internal knowledge than from the full text of the books.

To better compare the evaluator models, the scores for the summaries generated by GPT-4.1 and Gemini 2.5 Flash are shown as points in two dimensions, one corresponding to the score when Gemini 2.5 Flash is the judge and the other when GPT-4.1 is the evaluator. This visualizes the consistency in the evaluations. It can be seen that the evaluators generally agree on which summaries are better, with GPT-4.1 favoring internal summaries for the two generators and temperatures. Increasing the temperature slightly increases the dispersion of the results. 


\section{Conclusion}
\label{sec_conclusion}

In this paper, we have compared LLM-generated summaries of well-known books based on prior knowledge with those generated from the full text of the books. The results show that summaries produced with the full text are consistently preferred over summaries generated purely from parametric knowledge. However, for some books (e.g., The Divine Comedy, Hamlet, Ulysses, The Count of Monte Cristo) internal summaries outperform the ones done with the full text in many configurations. Therefore, in some cases, remembering is better than reading. This suggests that the summarization of long texts remains a challenging task to LLMs when dealing with very long, poetic, or structurally complex works. Nevertheless, further research is required to confirm whether this 'reading over remembering' effect persists across a broader range of models and literary works.

\section*{Limitations}

The findings of this work should be interpreted with caution, as several limitations may influence the results:

\begin{itemize}[nosep]
\item Models: only two closed commercial LLMs (GPT-4.1, Gemini-2.5-Flash) were tested; the results may change with versions, other vendors, or open-weight models.
\item Dataset: the 25 books evaluated are well-known, mostly Western books. The results may be different for other books.
\item Prompting: single prompts were evaluated for summary generation and evaluation; other prompt styles should be evaluated to assess their impact on the results.
\item Judge: LLM-as-a-judge can be biased toward its own generations or produce inconsistent evaluations. Although cross-judging was implemented, additional validation from human experts would be beneficial.
\item Generalization: Findings may not transfer to other documents such as technical documentation or less known books.
\end{itemize}

\section{Acknowledgements}

This work was supported by the Agencia Estatal de Investigación (AEI) (doi:10.13039/501100011033) under Grants FUN4DATE (PID2022-136684OB-C22) and SMARTY (PCI2024-153434), by TUCAN6-CM (TEC-2024/COM460) funded by CM (ORDEN 5696/2024) and by the European Commission through the Chips Act Joint Undertaking project SMARTY (Grant 101140087). The access to the models was provided by the OpenAI researcher access program and from Google.org and the Google Cloud Research Credits program for the Gemini Academic Program.

\bibliographystyle{unsrt}
\bibliography{Summaries}

\end{document}